\documentclass[10pt,twocolumn,letterpaper]{article}

\usepackage{wacv}              %

\usepackage{circledsteps}
\usepackage{graphicx}
\usepackage{amsmath}
\usepackage{amssymb}
\usepackage{booktabs}
\usepackage{array}
\usepackage{amsmath}
\usepackage{graphicx}
\usepackage{newtxmath}
\usepackage{caption}
\usepackage{subcaption}
\usepackage{colortbl}
\usepackage{xcolor}
\usepackage{paralist}
\usepackage{mathtools}
\usepackage{caption}
\usepackage{pifont}
\newcommand{\cmark}{\text{\ding{51}}}
\newcommand{\xmark}{\text{\ding{55}}}
\usepackage{paralist}
\usepackage{siunitx}

\definecolor{mygray}{RGB}{85,85,85}
\newcommand{\mycirc}[1]{\Circled[fill color=mygray, inner color=white, outer color=white]{#1}}

\newcolumntype{s}{>{\columncolor[gray]{.85}[.5\tabcolsep]}c}
\newcolumntype{d}{S[table-format=2.2]}

\definecolor{pltgreen}{RGB}{0,128,0}  %
\definecolor{methodblue}{RGB}{50,145,204}
\definecolor{methodyellow}{RGB}{248,187,47}
\definecolor{methodred}{RGB}{237,54,21}	
\definecolor{methodgreen}{RGB}{63,162,78}
\definecolor{methodpurple}{RGB}{93,50,145}

\definecolor{guideblue}{RGB}{50,145,204}
\definecolor{attackred}{RGB}{237,54,21}
\definecolor{renderpurple}{RGB}{93,50,145}

\newcommand{\xhdr}[1]{\vspace{3pt}\noindent\textbf{#1}}
\newcommand{\xhdrflat}[1]{\noindent\textbf{#1}}

\newcommand{\tabref}[1]{Tab.~\ref{#1}\xspace}
\newcommand{\figref}[1]{Fig.~\ref{#1}\xspace}
\newcommand{\secref}[1]{Sec.~\ref{#1}\xspace}

\DeclareMathOperator*{\argmax}{argmax}

\newcommand{\csection}[1]{
    \vspace{-2pt}
    \section{#1}
    \vspace{-4pt}
}

\newcommand{\csubsection}[1]{
    \vspace{-1pt}
    \subsection{#1}
    \vspace{-2pt}
}

\usepackage[pagebackref,breaklinks,colorlinks]{hyperref}

\usepackage[capitalize]{cleveref}
\crefname{section}{Sec.}{Secs.}
\Crefname{section}{Section}{Sections}
\Crefname{table}{Table}{Tables}
\crefname{table}{Tab.}{Tabs.}

\def\wacvPaperID{2300} %
\def\confName{WACV}
\def\confYear{2025}

\begin{document}

\title{Hijacking Vision-and-Language Navigation Agents \\ with Adversarial Environmental Attacks}

\author{Zijiao Yang, Xiangxi Shi, Eric Slyman, Stefan Lee\\
Oregon State University\\
{\tt\small \{yangziji,shixia,slymane,leestef\}@oregonstate.edu}
}
\maketitle

\begin{abstract}

Assistive embodied agents that can be instructed in natural language to perform tasks in open-world environments have the potential to significantly impact labor tasks like manufacturing or in-home care -- benefiting the lives of those who come to depend on them. In this work, we consider how this benefit might be hijacked by local modifications in the appearance of the agent's operating environment. Specifically, we take the popular Vision-and-Language Navigation (VLN) task as a representative setting and develop a whitebox adversarial attack that optimizes a 3D \emph{attack object}'s appearance to induce desired behaviors in pretrained VLN agents that observe it in the environment. We demonstrate that the proposed attack can cause VLN agents to ignore their instructions and execute alternative actions after encountering the attack object -- even for instructions and agent paths \emph{not} considered when optimizing the attack. For these novel settings, we find our attacks can induce early-termination behaviors or divert an agent along an attacker-defined multi-step trajectory. Under both conditions, environmental attacks significantly reduce agent capabilities to successfully follow user instructions.

\end{abstract}

\csection{Introduction}
\label{sec:intro}

Developing assistants that follow natural language instructions to execute complex tasks in open-world environments is a compelling task that enables applications like household robotics~\cite{abou2020systematic} and automated warehouse management~\cite{sodiya2024ai}. To study computational mechanisms underpinning the multimodal reasoning skills necessary to support such applications, the community has rallied around benchmark tasks like Vision-and-Language Navigation (VLN) that require agents to traverse an environment by grounding a natural language navigation instruction to visual observations \cite{anderson2018vision}. A key focus in VLN is that agents must generalize to new natural language instructions in novel environments. However, deploying a learning-based system in uncontrolled environments opens the possibility of malicious actors that modify the surroundings to intentionally impact the system's performance, reliability, or efficiency~\cite{barreno2006can}.

One class of such environmental modification is adversarial attacks that optimize for patterns or objects that, when photographed, induce errors in learning-based systems -- e.g., adversarial stickers \cite{eykholt2018robust}, t-shirts \cite{wu2020making}, or 3D objects \cite{athalye2017synthesizing}. Adversarial attacks directly optimizing digital inputs are more common and have been applied to image classifiers \cite{costa2024deep}, natural language processing models~\cite{morris2020textattack,yoo2021towards}, sequential decision making policies~\cite{huang2017adversarial}, and even multimodal vision language models (VLMs) \cite{zhao2023evaluate,luo2024imageworth1000lies}.

In this work, we study adversarial attacks for the VLN task -- a multimodal sequential decision making task requiring vision-and-language reasoning. We design a whitebox environmental adversarial attack paradigm that leverages differential rendering in 3D mesh environments to optimize the appearance of 3D \emph{attack objects} in VLN environments to induce specific behavior from a trained VLN agent. These attacks are optimized to either (1) force the VLN agent to immediately terminate the episode by issuing a stop command even if far from its instructed goal or (2) follow an attacker-defined trajectory to a location not specified by the original instruction. We study the effectiveness of these attacks in response to varying natural language instruction and navigation history.

To assess our method and explore model vulnerabilities in VLN, we study its effects on a representative VLN agent~\cite{chen2021history} in the popular R2R ~\cite{anderson2018vision} and RxR ~\cite{ku2020room} datasets. For both stopping and trajectory-following attacks, we find the presence of attacked objects results in significant disruption of VLN performance -- e.g., a reduction in success rate from 82.42\% to 53.85\% on a trajectory-following attacked subset of R2R. Moreover, we find varying degrees of success for inducing adversary-desired behaviors for novel instruction-trajectory pairs -- e.g., stop attacks causing episode termination in 75\% of cases and trajectory-following attacks causing agents to arrive at the new destination 20\% of the time in R2R. For instruction-trajectory pairs used during attack optimization, these rates increase substantially. Finally, we assess the influence of method hyperparameters and factors like object size and category on attack success.

\xhdr{Contributions.} Summarizing this work, we:
\begin{compactitem}[\hspace{3pt}--]
\item Develop an adversarial attack framework for controlling the trajectories of VLN agents that uses differentiable rendering to modify the appearance of 3D scene objects.
\item Demonstrate that the resulting attacks are effective at altering the behavior and performance of a representative VLN model \cite{chen2021history} when generalizing to new instruction-trajectory instances in the attacked scene on representative VLN datasets.
\item Present statistical analysis to better understand what factors influence the success of these attacks.
\end{compactitem}

\csection{Related Work}
\label{sec:related}

\xhdrflat{Vision-and-Language Navigation (VLN).}
VLN requires an embodied agent to traverse an environment by grounding natural language instructions to visual observations \cite{anderson2018vision}. In this work, we study HAMT~\cite{chen2021history} as a representative VLN agent with a history mechanism that preserves information from past visual inputs for future decision-making. We refer readers to a recent survey for additional background ~\cite{gu2022vision}.

\xhdr{Generalizable Adversarial Attacks.}
Adversarial Machine Learning (AML)~\cite{barreno2006can} is critical for exposing vulnerabilities in ML systems by identifying inputs that induce failure states or controllable behaviors. Adversarial attacks, a common form of AML, craft inputs that provoke specific responses from models. While early attacks focus on  RGB image perturbations that cause misclassifications in vision systems~\cite{DBLP:journals/corr/SzegedyZSBEGF13,moosavi_dezfooli2016universal}, the scope of these attacks has broadened to include other domains like natural language processing~\cite{morris2020textattack,yoo2021towards}, control policies~\cite{huang2017adversarial}, and more~\cite{costa2024deep}.

Prior work demonstrates the generalization of adversarial attacks beyond pathological fine-grained RGB perturbations, highlighting vulnerabilities in safety-critical systems. \cite{moosavi_dezfooli2016universal} demonstrate attacks that generalize across images, while \cite{huang2017adversarial} show that similar attacks can extend and generalize to neural network policies. %
\cite{luo2024imageworth1000lies} study transferable  attacks for vision-and-language models such that perturbing a single image would mislead all predictions given different textual prompts.
Notably, \cite{kurakin2017adversarial, jan2019connecting} study robustness of adversarial images to digital-to-physical transformations like printing, showing that adversarial attacks are generalizable to the real world. Our work extends the understanding of adversarial attacks by focusing on multimodal embodied agents in the VLN setting, showing that localized consistent appearance modifications on 3D objects can significantly disrupt agent behavior, hijacking agent's sequential decision making process. Unlike previous studies demonstrating generalization across images, models, and policies, our attacks must generalize across unseen natural language instructions and navigation trajectories.

\xhdr{Adversarial Attacks on Embodied Agents.}
While early work attacking reinforcement learning (RL) policies consider perturbing an agent's visual inputs when playing Atari games~\cite{huang2017adversarial,lin2017tactics,zhang2020robust}, recent work considers the more complex task of attacking embodied agents in 3D environments~\cite{qiaoben2022consistent}. Two works in particular are most similar to ours. \cite{zhang2022navigation} consider attacking VLN agents, but do so in the setting of federated learning by leveraging a malicious client to output poisoned data causing the global agent to disregard the natural language instruction and instead follow a series of visual triggers. \cite{liu2020spatiotemporal} study 3D adversarial attacks on EQA~\cite{embodiedqa}. Unlike our work, they consider synthetic (rather than natural) QA-like language instructions, attack the question-answering output of the agent rather than its navigation trajectory, and operate in unrealistic synthetic scenes.

\xhdr{Environment Attacks.} Adversarial attacks often target the environment where a digital image is captured rather than altering the image pixels directly. \cite{brown2017adversarial} develop ``adversarial patches'' that induce specific misclassifications when placed on images. Similarly, \cite{eykholt2018robust} create robust physical sticker-based attacks that mimic graffiti and remain effective under varied real-world conditions. Several works consider camouflage textures that can be applied to real objects. \cite{wu2020making} design adversarial shirt patterns to deceive object detection models. Similar methods are applied to varied objects~\cite{duan2020adversarial} and cars~\cite{wang2021dual} to fool image classifiers and object detectors, respectively. \cite{li2019adversarial} use physical semi-translucent stickers on camera lenses, attacking the camera apparatus instead of objects in the environment. Several recent works leverage differentiable rendering to perturb 3D scenes. \cite{athalye2017synthesizing,pestana2022transferable} modify the texture of 3D objects to induce robust 2D adversarial image attacks, while \cite{miao2022isometric,zhang20213d} focus on attacks against models which operate in 3D environments themselves. %

In contrast, our work studies environmental adversarial attacks on VLN agents. This approach involves permuting the RGB texture of an object in the scene during navigation, directing the agent to follow a different path from the one specified in the natural language instruction. Unlike prior work primarily targeting static image classification and detection, this work focuses on dynamic navigation tasks and the interplay between visual and linguistic inputs.

\csection{Adversarial Attacks on VLN Agents}

\label{sec:method}

\begin{figure*}[t]
    \centering
    \includegraphics[width=0.925\textwidth]{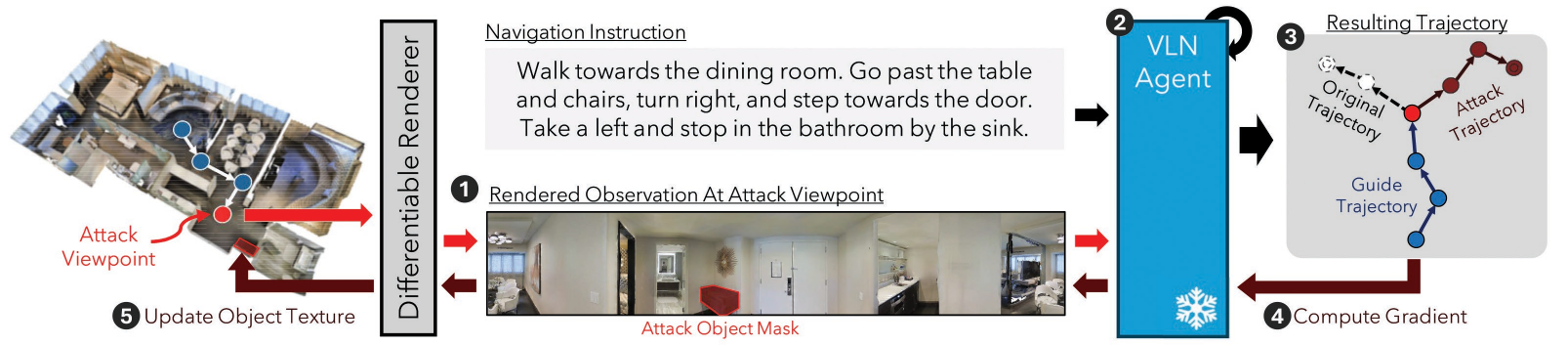}
    \caption{We directly optimize the appearance of an in-environment object to control the trajectory of a trained VLN agent using a differentiable renderer. \mycirc{1} Adversarial observations are rendered at an attack viewpoint containing the attack object. \mycirc{2} The VLN agent takes this observation as input and \mycirc{3} we supervise the agent's trajectory from this point to match a predetermined attack trajectory. \mycirc{4} We compute loss gradients with respect to the object texture and use them to \mycirc{5} update the object's appearance in the 3D mesh. }
    \label{fig:overview}
\end{figure*}

Before describing our adversarial environmental attack framework, we establish notation for the task setting.

\xhdr{VLN Episodes and Agents.} A single Vision-and-Language Navigation (VLN) episode can be defined as a tuple ($\mathcal{E}$, $I$, $\tau$) consisting of an environment $\mathcal{E}$ in which the agent is situated, a natural language navigation instruction $I$, and a trajectory $\tau$ through $\mathcal{E}$ corresponding to accurately following $I$. 
In standard discrete VLN settings \cite{anderson2018vision}, the environment $\mathcal{E}$ is represented as an undirected navigation graph with nodes $\mathcal{V_{\mathcal{E}}}$ corresponding to a discrete set of environment locations the agent can visit and edges $L_{\mathcal{E}}$ between nodes which the agent is allowed to traverse. A trajectory is then a sequence $\tau=[v_0,v_1, ..., v_n]$ where all $v_t\in\mathcal{V}_\mathcal{E}$ and $(v_{t-1},v_t) \in L_\mathcal{E}$. Further, each location $v_i$ is associated with a panoramic observation of $\mathcal{E}$ taken at $v_i$ which we denote as $o_{v_i}$.

Successfully following an instruction in a new environment requires making a sequence of accurate predictions about where to navigate next. Given an instruction $I$, a partial trajectory $[v_0, ..., v_t]$ up to the current step $t$, and the corresponding observations $[o_{v_0}, ..., o_{v_t}]$, a VLN agent $\pi_\theta$ produces a 
distribution over candidate navigation actions $\mathcal{A}(v_t) = N(v_t) \cup \{\mbox{\texttt{STOP}}\}$, where $N(v_t)$ is the set of nodes connected to $v_t$ and the \texttt{STOP} action terminates the episode. The agent's action is then selected as
\begin{align*}
    \argmax_{a \in \mathcal{A}(v_t)} ~~\pi_\theta\big(~a\mid I, ~[v_0, ..., v_t]~, ~[o_{v_0}, ..., o_{v_t}]~\big).
\end{align*}%
The trajectory terminates when the agent issues the \texttt{STOP} command and agent performance is based on the similarity of the resulting trajectory to the ground truth $\tau$.

\xhdr{Adversarial Attacks on VLN Agents.} As the VLN task is a proxy for real language-guided embodied systems, we consider how an adversary might gain influence over a real VLN agent's behavior when it is deployed in uncontrolled environments. For instance, how a delivery robot might be diverted to a desired location or a surveillance drone be made to end its rounds prematurely -- regardless of the original instructions they were given by their users. In both, an adversary would like to be able to dictate the sequence of actions an agent takes after encountering the attack. %

In this embodied setting, the agent's visual observations would be acquired using an onboard camera. Consequentially, standard adversarial attacks that directly modify image pixels \cite{moosavi_dezfooli2016universal} would only be possible if an adversary has gained software access to the agent. Alternatively, transparent film-based attacks that are applied over a physical camera lens have been developed \cite{li2019adversarial} that might be applied, but assume physical access to the agent a priori. 

Rather than assuming access to individual agents, we consider a class of object-based environmental attacks where an adversary has control over the appearance of an \emph{attack object} in the 3D environment. Further, we do not assume the adversary knows the particular instruction given to the agent or the partial trajectory it has taken before encountering the attack object. That is to say, a successful attack must generalize to new language instructions and observation histories. While many goals for the adversary are possible, we consider \emph{trajectory} attacks where an attacked agent is made to follow an adversary-defined trajectory after encountering the attack object. This requires the attack to influence a whole sequence of decisions, even when the attack object itself may no longer be visible.

Formalizing this slightly, we seek a procedure to adjust an object's appearance in order to maximize the expected likelihood of a given VLN agent following an attack trajectory after entering the attack object's location. To promote generalization of the attack, this expectation is over pairs of natural instructions and corresponding agent trajectories prior to encountering the attack object. We examine this in the whitebox setting where the adversary also has access to the VLN model structure and weights being deployed as well as the target environment.

\csubsection{Attack Methodology}

Our attack framework is demonstrated in \figref{fig:overview} for a single trajectory. Given a textured 3D mesh representation of the environment, we select a viewpoint and object to attack. \mycirc{1}  For an agent approaching this \emph{attack viewpoint}, we use a differentiable rendering pipeline to produce the corresponding panoramic observation. \mycirc{2} Given the history of its trajectory so far, the instruction, and the rendered adversarial observation, the VLN agent produces a distribution over next actions. \mycirc{3} We supervise the agent to assign high probability to the attack trajectory and \mycirc{4} backpropagate gradients of this loss all the way to the object texture in the 3D mesh. \mycirc{5} With this gradient, we update the object's texture while respecting attack magnitude constraints. To provide generalization, we perform this optimization over a set of episodes that pass through the attack viewpoint. 

\xhdr{Attack Viewpoint and Trajectory.}  %
We denote the viewpoint where the attack object is located as the attack viewpoint $v_{\text{ATK}}$. After entering the attack viewpoint, a successfully attacked VLN agent would then follow the attack trajectory $\hat{\tau} = [\hat{v}_0, \hat{v}_1, ..., \hat{v}_L, \texttt{STOP}]$ terminating at the target location by issuing the \texttt{STOP} command. We refer to the agent's trajectory up to encountering $v_{\text{ATK}}$ as the \emph{guide trajectory} and denote it as $\tau_{\leq v_{ {\text{ATK}}}}$. We denote the corresponding observation sequence in a similar fashion as $o_{\leq v_{{\text{ATK}}}}$.

\xhdr{Observation Rendering.} Given a 3D model of the environment $\mathcal{E}$ consisting of a 3D mesh denoted $M_{\mathcal{E}}$ and corresponding texture atlas image $T_{\mathcal{E}}$, we apply the  \textsc{pytorch3D} \cite{ravi2020accelerating} differentiable renderer to produce observations through which gradients can affect object textures. In contrast to image-level attacks, our framework takes into account the view consistency of the attacked object by directly optimizing an explicit 3D representation.%

As differentiable rendering has a high memory and computational cost, we limit its use to a subset of viewpoints along the attack trajectory. Specifically, we render observations for the attack viewpoint and the following next two steps. This allows the object's appearance to not just affect the decision where the object is most visible, but also future decisions more effectively~\cite{qiu2022adversarial,roth2021token} while ensuring consistent appearance  across these different views. %
To further reduce costs, we only render panorama sub-images that contain the attack object. When attacking an object, we produce an object mask $D$ such that $D_{i}$ is 1 if the $i^{th}$ face in the mesh is used to render the object and 0 otherwise. 

\xhdr{Optimization.} As we would like our attacks to generalize to new trajectories and instructions, we consider a set of VLN episodes  $\{(\mathcal{E}, I^{(i)}, \tau^{(i)})\}_{i=1}^n$ for training the attack such that each ground truth trajectory $\tau_i$ includes $v_{\texttt{ATK}}$. To maximize the expected likelihood described above, we consider the cross entropy loss for the agent following the attack trajectory after arriving at the attack viewpoint $v_{\mbox{\texttt{ATK}}}$, denoting it as $\ell^{(i)}_{\mbox{CE}}(T_\mathcal{E})$ equal to
\newcommand{\vadj}{\vphantom{ a_{a_a} }}
\begin{multline*}
-\sum_{t=0}^{L+1} \ln \pi_\theta\Big( \hat{v}_t ~|~ I^{(i)},  [~\textcolor{guideblue}{\underbrace{\vadj \tau^{(i)}_{\leq v_{{\text{ATK}}}}}_{\makebox[10pt]{\footnotesize\shortstack{Guide\\Traj.}}}},
    \textcolor{attackred}{\underbrace{\vadj \hat{\tau}_{<t}}_{\makebox[10pt]{\footnotesize\shortstack{Attack\\Traj.}}}}~], [~\underbrace{\vadj o^{(i)}_{\leq v_{\text{ATK}}}, o_{\hat{v}_{<t}}}_{\makebox[0pt]{\footnotesize\shortstack{Panoramic\\Observations}}}~]\Big)
    \label{eq:loss}
\end{multline*}%
where the summation traverses each step in the attack trajectory and $\hat{\tau}_{<t}$ and $o_{\hat{v}_{<t}}$ denote the attack trajectory and observations up to time step $t$. We denote the aggregation of this loss across a batch of attack training episodes as $\mathcal{L}_{\mbox{CE}}$.

To update object appearance, we take the gradient of $\mathcal{L}_{\mbox{CE}}$ with respect to the object texture, computing $D\odot\nabla_{T_{\mathcal{E}}}\mathcal{L}_{\mbox{CE}}$ where $\odot$ denotes a gradient masking for non-object mesh faces and $T_{\mathcal{E}}$ is the texture atlas image. With this, we use the ADAM optimizer \cite{kingma2014adam} to update $T_\mathcal{E}$. In line with existing Projected Gradient Descent-based attack strategies \cite{madry2018towards}, we limit the $L_\infty$ norm of the changes from the original to some attack magnitude constant $\epsilon$ and clamp color values to [0,1]. This iterates over multiple batches and update steps. The final output of this optimization process is an altered 3D environment model 
$\mathcal{E}'$ in which the appearance of the attack object has been modified. We can then evaluate the attack on new instruction-trajectory pairs.

\csubsection{Generating and Evaluating Attacks}%
\label{sec:gen_attacks}
To evaluate our attack strategy, we consider  VLN episodes from the Room-2-Room (R2R) dataset \cite{anderson2018vision} consisting of nearly 22,000 episodes across 90 indoor environments from the MP3D dataset \cite{chang2017matterport3d}. These episodes are divided into \texttt{R2R-train}, \texttt{R2R-val-seen}, \texttt{R2R-val-unseen}, and \texttt{R2R-test}. For our purposes, we focus on \texttt{R2R-train} and \texttt{R2R-val-seen} which share the same set of environments but different trajectory-instruction pairs. We consider episodes in \texttt{R2R-val-seen} to select attack viewpoints and objects that have sufficient supporting episodes in \texttt{R2R-train} to train an attack. Note that while VLN agents have been trained on \texttt{R2R-train}, \texttt{R2R-val-seen} episodes are novel instruction-trajectory pairs to the model and attacks. 

We also consider RxR~\cite{ku2020room} to additionally verify effectiveness of our method. RxR is a multilingual and larger VLN dataset consists of more than 16,000 paths and 120,000 fine-grained grounded instructions. RxR situated in the same set of house environment as R2R and has longer instruction and longer path on average: 78 \vs 29 words, 8 vs 5 edges. Similar as in R2R, we consider \texttt{RxR-train-guide} and \texttt{RxR-val-seen-guide} that share the same set of environments, and use \texttt{RxR-train-guide} to train attack while using \texttt{RxR-val-seen-guide} to select attack viewpoints and objects. 

We use \texttt{train-data} and \texttt{val-seen-data} for ease of description on data processing over R2R and RxR.

\xhdr{Candidate Attack Objects.} To select attack objects, we consider object visibility annotations from \cite{qi2020reverie} associated with 3D object segmentations from MP3D \cite{chang2017matterport3d} -- examining objects from the most common categories of \texttt{chair}, \texttt{cabinet}, \texttt{table}, \texttt{plant}, \texttt{sofa}, and \texttt{TV monitor}. For each viewpoint in R2R/RxR, we compile a list of these objects visible from that location. Further, we discard any object that does not make up at least 40\% of the pixels in at least one of the 36 panoramic sub-images for this viewpoint. To compute this visibility score, we render the object mask at each of the sub-images and compute the pixel ratio.

\xhdr{Constructing Attack Instances.} For each episode ($\mathcal{E}$, $I$, $\tau$) in \texttt{val-seen-data}, we check if there exists any viewpoints $v{\in}\tau$ with a valid candidate attack object $X$ and at least 5 episodes in \texttt{train-data} that pass through $v$. Further, we exclude any training episode if its guide trajectory matches that of the val-seen episode to ensure our evaluation represents a novel observation history. If multiple viewpoints meet this criteria, we select the one corresponding to the object with the largest visibility score and denote it as the attack viewpoint $v_{\mbox{\texttt{ATK}}}$ for this instance. Finally, we generate an attack trajectory by selecting a viewpoint that is at least three meters away from the original final viewpoint $v_K$ in $\tau$ and find a shortest path $\tau_{\mbox{\texttt{ATK}}}=v_{\mbox{\texttt{ATK}}}, ..., v_K$ to it from the attack viewpoint. A resulting attack instance $j$ then consists of an environment $\mathcal{E}_j$, instruction $I_j$, ground truth trajectory $\tau_j$, attack object $X_j$, attack viewpoint $v_{\mbox{\texttt{ATK}}_j}$, attack trajectory $\tau_{\mbox{\texttt{ATK}}_j}$, and set of training instances $\mathcal{D}_j=\{\mathcal{E}_j, I^{(i)}, \tau^{(i)}\}_{i=1}^N$ $\subset$ \texttt{train-data}. When optimizing our attack, we split $\mathcal{D}_j$ 80/20 into $\mathcal{D}_{j}^{\mathtt{Train}}$ and $\mathcal{D}_{j}^{\mathtt{Val}}$ to produce a validation set. In total, this process yields a test set with 273 attack instances and 254 attack instances for R2R and RxR respectively. Notice we sample a subset of attack instances constructed from RxR to accommodate compute resource and time limit.

\xhdr{Evaluation.} For each attack instance, we run our optimization procedure described in the preceding subsection, yielding an attacked environment $\mathcal{E}'$. During evaluation, we render all observations using $\mathcal{E}'$ rather than just near the attack viewpoint. To evaluate the effect on the attack instance, we force the VLN agent through the guide trajectory until reaching $v_{\mbox{\texttt{ATK}}}$ and then allow it to autoregessively navigate the environment until it terminates by issuing the \texttt{STOP} action. We consider two evaluation strategies. In the first, we measure how the attack affects the agent's ability to reach the goal specified by the instruction. For the second, we measure whether the attacked environment actually causes the agent to follow the corresponding attack trajectory. For both, we use standard VLN path comparison and success metrics described in the following section -- comparing the agents trajectory after $v_{\mbox{\texttt{ATK}}}$ with either the rest of the original trajectory $\tau$ or the attack trajectory $\tau_{\mbox{\texttt{ATK}}}$.

\begin{table}[t]
\centering
\begin{center}
\setlength{\tabcolsep}{5pt}
\renewcommand{\arraystretch}{0}
\begin{tabular}{c c c}
\frame{\includegraphics[height=0.55in, width=0.9in]{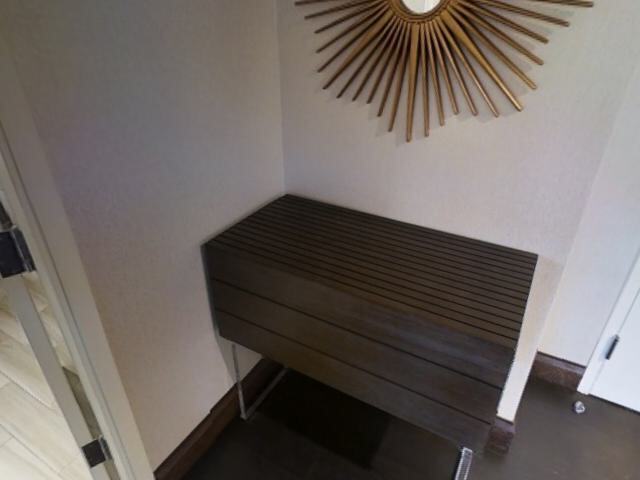}} &
\frame{\includegraphics[height=0.55in, width=0.9in]{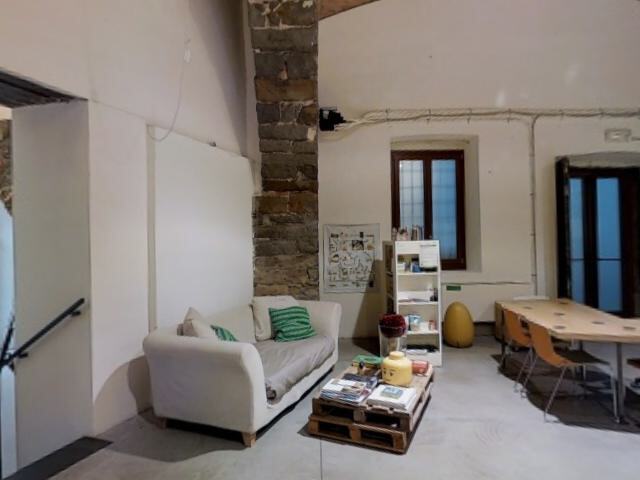}} &
\frame{\includegraphics[height=0.55in, width=0.9in]{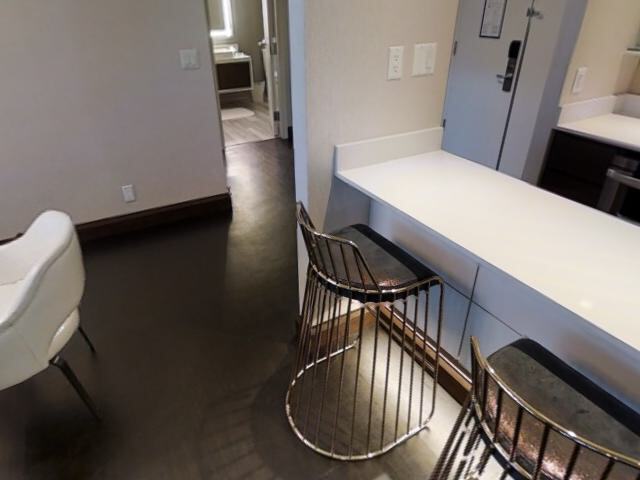}}\\
\frame{\includegraphics[height=0.55in, width=0.9in]{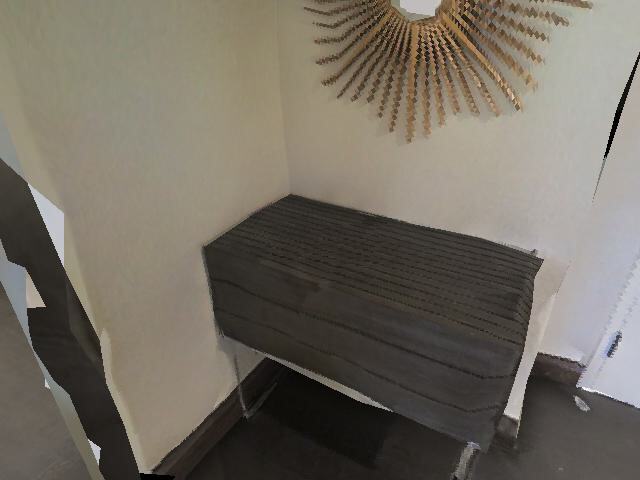}} &
\frame{\includegraphics[height=0.55in, width=0.9in]{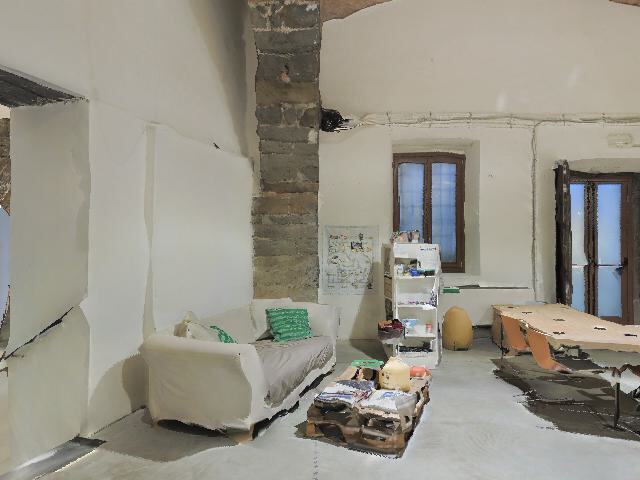}} &
\frame{\includegraphics[height=0.55in, width=0.9in]{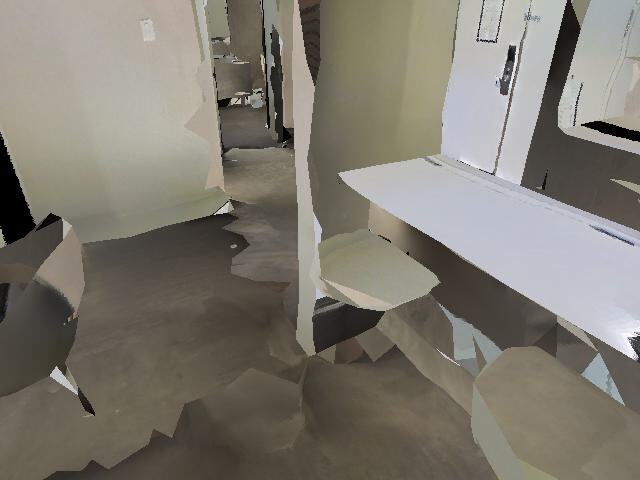}}\\[5pt]
\end{tabular}%
\hspace{2pt}
\renewcommand{\arraystretch}{1}
\setlength{\tabcolsep}{6pt}
 \resizebox{0.91\columnwidth}{!}{%
\begin{tabular}{l p{0.5em} cc p{0.5em} cc p{0.4em}}
\toprule
 && \multicolumn{2}{c}{R2R} && \multicolumn{2}{c}{RxR} \\ \cmidrule(lr){3-4} \cmidrule(lr){6-7}

\multicolumn{1}{l}{Metrics}  && Real & Rendered && Real & Rendered  \\ \midrule

SR \textuparrow     && 75.61 & 69.64 && 59.37  & 53.00 \\
nDTW \textuparrow   && 78.33 & 72.16 && 65.41 & 59.13\\ 
\bottomrule
\end{tabular}%
 }
\caption{\emph{Top:} Real scene images (top row) are higher fidelity than rendered images (bottom row). \emph{Bottom:} Despite this, we find VLN performance for HAMT on the \texttt{val-seen} split for R2R and RxR only drop by marginal amounts when replacing real images with rendered ones.}
\label{tab:mesh-effect on vln}
\end{center}
\vspace{-15pt}
\end{table}

\begin{table*}[t]
\centering
\begin{subtable}[t]{0.49\textwidth}
\centering
\renewcommand*{\arraystretch}{1.5}
\setlength{\tabcolsep}{3pt}
\resizebox{\columnwidth}{!}{
\begin{tabular}{l p{0em} ccs p{0.15em} ccs p{0.15em} ccs p{0em}}
\toprule
 \shortstack{\textcolor{darkgray}{\textbf{R2R}}\\\vphantom{\scriptsize a}}&& \multicolumn{3}{c}{\shortstack{Train\\ \scriptsize{($\subset$ \texttt{R2R train})}}} & &\multicolumn{3}{c}{\shortstack{Validation\\ \scriptsize{($\subset$ \texttt{R2R train})}}} && \multicolumn{3}{c}{\shortstack{Test\\ \scriptsize{($\subset$ \texttt{R2R val-seen})}}} & \\ \cmidrule(lr){3-5} \cmidrule(lr){7-9} \cmidrule(lr){11-13}

\multicolumn{2}{r}{Attacked:}        & {\cmark} & {\xmark} & {\large$\mathbf{\Delta}$} && {\cmark} & {\xmark} & {\large$\mathbf{\Delta}$} && {\cmark} & {\xmark} & {\large$\mathbf{\Delta}$} \\ \midrule

Stop \% \textuparrow      &&  98.28 & 1.80 & 96.48 && 80.70 & 1.97 & 78.73 && 75.98 & 0.98 & 75.00 \\
\bottomrule
\end{tabular}}
\end{subtable}
\hfill
\begin{subtable}[t]{0.49\textwidth}
\centering
\renewcommand*{\arraystretch}{1.6}
\setlength{\tabcolsep}{3pt}
\resizebox{\columnwidth}{!}{
\begin{tabular}{l p{0em} ccs p{0.15em} ccs p{0.15em} ccs p{0em}}
\toprule
 \shortstack{\textcolor{darkgray}{\textbf{RxR}}\\\vphantom{\scriptsize a}}&& \multicolumn{3}{c}{\shortstack{Train\\ \scriptsize{($\subset$ \texttt{RxR train})}}} & &\multicolumn{3}{c}{\shortstack{Validation\\ \scriptsize{($\subset$ \texttt{RxR train})}}} && \multicolumn{3}{c}{\shortstack{Test\\ \scriptsize{($\subset$ \texttt{RxR val-seen})}}} & \\ \cmidrule(lr){3-5} \cmidrule(lr){7-9} \cmidrule(lr){11-13}

\multicolumn{2}{r}{Attacked:}        & {\cmark} & {\xmark} & {\large$\mathbf{\Delta}$} && {\cmark} & {\xmark} & {\large$\mathbf{\Delta}$} && {\cmark} & {\xmark} & {\large$\mathbf{\Delta}$} \\ \midrule

Stop \% \textuparrow      &&  46.05 & 13.68 & 32.37 && 33.04 & 13.91 & 19.13 && 25.20 & 14.17 & 11.03 \\
\bottomrule
\end{tabular}}%
\end{subtable}

\caption{\emph{Single-step Attacks for R2R (left) and RxR (right).} %
We compare stop rate for the HAMT agent operating in attacked (\cmark) and unaltered (\xmark) environments as well as the difference ($\mathbf{\Delta}$). Across all settings, we see our attacks substantially increase the likelihood of a VLN agent terminating prematurely -- especially for R2R where 75\% of novel trajectory-instruction pairs are ended by the agent.}
\label{tab:singlepoint}
\vspace{-6pt}
\end{table*}

\csection{Experimental Results}
\label{sec:experiments}
\label{sec:results}
\xhdrflat{Metrics.} For evaluating attack effectiveness, we focus on two primary VLN metrics \cite{ilharco2019general} -- success rate (SR) and normalized dynamic time warping (nDTW). Given a predicted trajectory and a reference trajectory, success requires the final viewpoint in the predicted trajectory be no further than 3m from that of the reference. Meanwhile, nDTW is a path similarity metric which is maximized if paths match exactly. %

\xhdr{VLN Model.} We conduct our attack experiment using the HAMT~\cite{chen2021history} model. HAMT is a widely-used  VLN model in the literature and representative of common architectures. Visual inputs are encoded using a ViT \cite{dosovitskiy2020image} image encoder. Trajectory history is encoded in a hierarchical fashion then jointly combined with encodings of instruction text and the current observation to predict the next action. It is trained in multiple stages with auxiliary losses followed by finetuning on VLN tasks with supervised and reinforcement learning. We use the best performing model weights provided by authors and keep weights frozen at all times.

\xhdr{Implementation Details.} For attack optimization, we set ADAM's learning rate to 1e-2 and exponential moving average rates to $\beta_1=0.9$ and $\beta_2=0.999$. For attack magnitude, we set $\epsilon=0.3$. To further reduce computational and memory costs, we decimate the environment meshes from MP3D and use lower-res texture atlas images. Optimization runs for 300 batches (iterations) with batch size 16, and saves checkpoints every 30 iterations. For RxR dataset, we increase to 600 batches to help convergence. After training, the checkpoint with highest nDTW with respect to attack trajectory $v_{\mbox{\texttt{ATK}}_j}$ on the validation set $\mathcal{D}_j$ of each attack instance j is retained. Recall that each attack instance is trained independently. Each attack training was performed on an NVIDIA A40 GPU and took on average 40 minutes for R2R and 90 minutes for RxR per attack.

\xhdr{Quantifying Domain Gap from Rendering.}
Our approach relies on rendering images from reconstructed MP3D environments; however, these reconstructions can be noisy and we further reduce their fidelity to accommodate available compute. As a result, the rendered images are lower quality than those with which VLN agents are typically trained -- especially for scenes containing thin or reflective structures. If this alone significantly affects VLN agent performance, we need to account for it when evaluating the impact of rendered attacks. %
To evaluate this gap, we compare performance of HAMT on the original R2R/RxR panoramas and those rendered from unattacked 3D reconstructions of the same environments. As shown in \tabref{tab:mesh-effect on vln}, the experiment using rendered imagery does demonstrate some reduced performance -- approximately 6-7 points in success or path following metrics -- but the model retains the bulk of its capabilities. To be considered successful, our attacks will need to illicit larger reductions than these.

\xhdr{Evaluation.} We evaluate in the following conditions: 

\noindent -- {\texttt{Train}.} While our primary focus is on attacks that generalize to new trajectory-instruction pairs, evaluating attack success for the examples on which they are optimized is a common setting for adversarial examples \cite{costa2024deep}. We evaluate each attack instance on the examples used to optimize the attack -- i.e., each episode in $\mathcal{D}_{j}^{\mathtt{Train}}$ is evaluated in the attacked environment $\mathcal{E}'_j$. Metrics are aggregated across all training episodes from all attack instances. Note that these instances belong to the \texttt{R2R-train} / \texttt{RxR-train-guide} set and thus have been seen by the HAMT model during training. As such, they require \emph{no} generalization from the attack \emph{or} the VLN model. %
   
\noindent -- {\texttt{Validation}.} Analogously, episodes from each validation set $\mathcal{D}_{j}^{\mathtt{Val}}$ are evaluated on their corresponding attacked environment $\mathcal{E}'_j$ and metrics are aggregated across all attack instances. As these episodes are also from \texttt{R2R-train} / \texttt{RxR-train-guide}, they are only novel to the attack and not the model.  
    Note that checkpoint selection uses these episodes so attack performance is likely overestimated here.

\noindent -- {\texttt{Test}.} Finally, we evaluate each attack instance $I_j,  \tau_j$ taken from \texttt{R2R-val-seen} / \texttt{RxR-val-seen-guide} on the corresponding attacked environment $\mathcal{E}'_j$. These trajectory-instruction pairs have not been seen when optimizing the attack or training the VLN model.
    This introduces potential out-of-distribution signals in the text and history features, leading to unpredictable behavior when these inputs are processed by the model in the attacked environment $\mathcal{E}'_j$. %
    Therefore, we consider this to be the most challenging evaluation settings.%

\csubsection{Single-step Attacks}
Before reporting results for our trajectory-level attacks, we examine a special case of our attack framework -- producing attacks that cause agents to stop immediately. By setting the attack trajectory to be empty ($\tau_{\mbox{\texttt{ATK}}} = \{\}$) agents are supervised to issue the \texttt{STOP} action and terminate the episode at the attack viewpoint. All other details of our methodology remain as described. This setting is analogous to environmental attacks against image classification systems; however, we consider a VLN model which is also conditioned on language and navigation history. Additional filtering is applied during attack generation to ignore episodes which already terminate at $v_{\mbox{\texttt{ATK}}}$. 
We show stop rates in \tabref{tab:singlepoint} for the HAMT agent operating in attacked (\cmark) and unaltered (\xmark) environments. Note that unaltered (\xmark) environments are rendered with \textsc{pytorch3D} as in \tabref{tab:mesh-effect on vln}.

For R2R (left), we find this simple attack is highly effective and would be extremely disruptive. %
The agent stops immediately in 75.98\% of \texttt{Test} instances -- a 76x increase from the unaltered environments. For \texttt{Train} and \texttt{Validation} samples, stop rates approach 98.28\% and 80.70\% respectively. If deployable, these attacks could make no-go zones for VLN agents where the stop action is taken regardless of the user's original instruction.

For RxR (right), we also see increased stop rates; however, the magnitude of these effects are significantly smaller than in R2R. We believe this difference is due to RxR having substantially longer path lengths -- roughly $\sim$6.1 vs.~$\sim$10.5 average steps -- inducing strong biases in the trained VLN models. Our attack trajectories require an agent to stop between step 4 and 6 on average, requiring attacks to override a strong prior from model training.

\begin{table*}[t]
\centering
\begin{subtable}[t]{0.49\textwidth}
\centering
\renewcommand*{\arraystretch}{1.25}
\setlength{\tabcolsep}{3pt}
\resizebox{\columnwidth}{!}{
\begin{tabular}{l p{0em} ccs p{0.15em} ccs p{0.15em} ccs p{0em}}
\toprule
 \textbf{\shortstack{\textcolor{darkgray}{\textbf{R2R}}\\\vphantom{\scriptsize a}}}&& \multicolumn{3}{c}{\shortstack{Train\\ \scriptsize{($\subset$ \texttt{R2R train})}}} & &\multicolumn{3}{c}{\shortstack{Validation\\ \scriptsize{($\subset$ \texttt{R2R train})}}} && \multicolumn{3}{c}{\shortstack{Test\\ \scriptsize{($\subset$ \texttt{R2R val-seen})}}} & \\ \cmidrule(lr){3-5} \cmidrule(lr){7-9} \cmidrule(lr){11-13}

\multicolumn{2}{r}{Attacked:}        & {\cmark} & {\xmark} & {\large$\mathbf{\Delta}$} && {\cmark} & {\xmark} & {\large$\mathbf{\Delta}$} && {\cmark} & {\xmark} & {\large$\mathbf{\Delta}$} \\ \midrule

SR \textuparrow      && 30.26 & 1.15 & 29.11 && 22.01 & 1.04 & 20.97 && 21.61 & 3.66 & 17.95 \\
OSR \textuparrow      && 43.00 & 6.17 & 36.83 && 34.32 & 6.93 & 27.39 && 38.46 & 11.36 & 27.10 \\
nDTW \textuparrow   && 53.38 & 29.11 & 24.27 && 47.48 & 29.34 & 18.14 && 46.65 & 30.60 & 16.05 \\ \bottomrule
\end{tabular}%
}
\end{subtable}
\hfill
\begin{subtable}[t]{0.49\textwidth}
\centering
\centering
\renewcommand*{\arraystretch}{1.25}
\setlength{\tabcolsep}{3pt}
\resizebox{\columnwidth}{!}{
\begin{tabular}{l p{0em} ccs p{0.15em} ccs p{0.15em} ccs p{0em}}
\toprule
 \shortstack{\textcolor{darkgray}{\textbf{RxR}}\\\vphantom{\scriptsize a}}&& \multicolumn{3}{c}{\shortstack{Train\\ \scriptsize{($\subset$ \texttt{RxR train})}}} & &\multicolumn{3}{c}{\shortstack{Validation\\ \scriptsize{($\subset$ \texttt{RxR train})}}} && \multicolumn{3}{c}{\shortstack{Test\\ \scriptsize{($\subset$ \texttt{RxR val-seen})}}} & \\ \cmidrule(lr){3-5} \cmidrule(lr){7-9} \cmidrule(lr){11-13}

\multicolumn{2}{r}{Attacked:}        & {\cmark} & {\xmark} & {\large$\mathbf{\Delta}$} && {\cmark} & {\xmark} & {\large$\mathbf{\Delta}$} && {\cmark} & {\xmark} & {\large$\mathbf{\Delta}$} \\ \midrule

SR \textuparrow      && 25.26 & 5.30 & 19.95 && 16.23 & 2.61 & 13.62 && 20.47 & 5.51 & 14.96 \\
OSR \textuparrow   && 69.74 & 23.45 & 46.29 && 59.13 & 18.55 & 40.58 && 78.74 & 35.43 & 43.31 \\
nDTW \textuparrow   && 30.64 & 18.26 & 12.38 && 25.56 & 15.05 & 10.51 && 21.25 & 14.44 & 6.80 \\
\bottomrule
\end{tabular}%
}
\end{subtable}
\caption{\emph{Trajectory-level Attacks for R2R (left) and RxR (right).} We report results for trajectory-level attacks that cause agents to follow specific attack trajectories. We compare success rate (SR), oracle success rate (OSR), and normalized Dynamic Time Warping (nDTW) for the HAMT agent operating in attacked (\cmark) and unaltered (\xmark) environments as well as the difference ($\mathbf{\Delta}$). Higher values imply better adherence to the attack trajectory. Across all settings, our attacks increase the agent's likelihood of following the attack trajectory.}
\label{tab:attack_to_loc}

\end{table*}

\begin{figure}[t]
    \begin{center}
    \setlength{\tabcolsep}{1pt}
    \resizebox{\columnwidth}{!}{
    \begin{tabular}{c  c  c}
    \frame{\includegraphics[height=0.31in, width=0.55in]{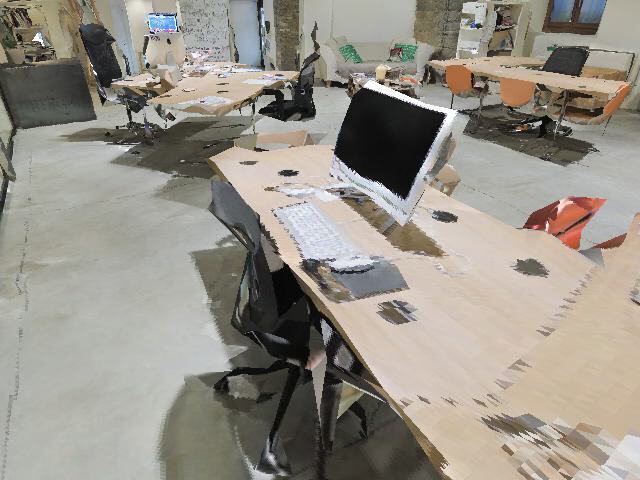}}  & 
    \frame{\includegraphics[height=0.31in, width=0.55in]{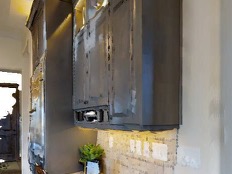}}  &
    \frame{\includegraphics[height=0.31in, width=0.55in]{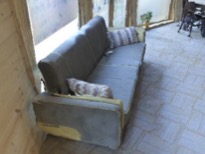}}  \\[-3pt]
    \frame{\includegraphics[height=0.31in, width=0.55in]{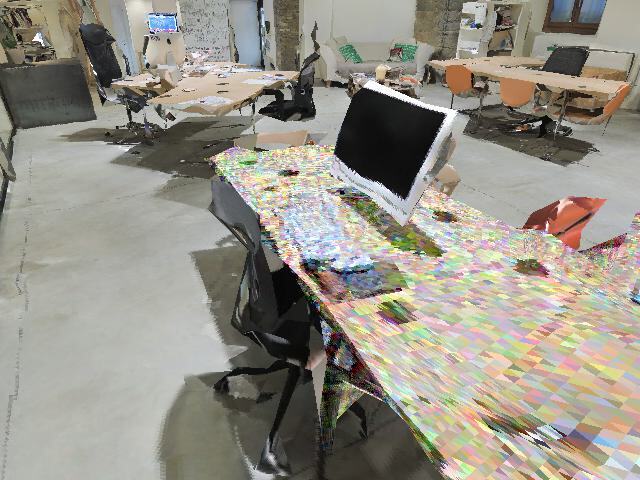}}  &
    \frame{\includegraphics[height=0.31in, width=0.55in]{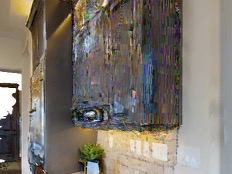}}  &
    \frame{\includegraphics[height=0.31in, width=0.55in]{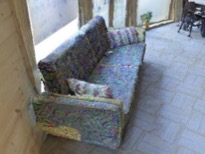}} 
    \end{tabular}}
    \end{center}
    \vspace{-20pt}
    \caption{Example original and attacked objects for a desk (left), cabinet (middle), and sofa (right) from trajectory-level attacks.}
    \label{fig:examples}
\vspace{-11pt}
\end{figure}

\begin{table*}[t]
\centering
\begin{subtable}[t]{0.49\textwidth}
\centering
\renewcommand*{\arraystretch}{1.25}
\setlength{\tabcolsep}{3pt}
\resizebox{\columnwidth}{!}{
\begin{tabular}{l p{0em} ccs p{0.15em} ccs p{0.15em} ccs p{0em}}
\toprule
 \shortstack{\textcolor{darkgray}{\textbf{R2R}}\\\vphantom{\scriptsize a}}&& \multicolumn{3}{c}{\shortstack{Train\\ \scriptsize{($\subset$ \texttt{R2R train})}}} & &\multicolumn{3}{c}{\shortstack{Validation\\ \scriptsize{($\subset$ \texttt{R2R train})}}} && \multicolumn{3}{c}{\shortstack{Test\\ \scriptsize{($\subset$ \texttt{R2R val-seen})}}} & \\ \cmidrule(lr){3-5} \cmidrule(lr){7-9} \cmidrule(lr){11-13}
\multicolumn{2}{r}{Attacked:}        & {\cmark} & {\xmark} & {\large$\mathbf{\Delta}$} && {\cmark} & {\xmark} & {\large$\mathbf{\Delta}$} && {\cmark} & {\xmark} & {\large$\mathbf{\Delta}$} \\ \midrule

SR \textuparrow     && 46.11 & 91.35 & -45.24 && 55.46 & 91.33 & -35.87 && 53.85 & 82.42 & -28.57 \\
nDTW \textuparrow   && 47.44 & 87.73 & -40.29 && 58.21 & 87.06 & -28.85 && 53.68 & 81.18 & -27.50 \\ 

\bottomrule
\end{tabular}}%
\end{subtable}
\hfill
\begin{subtable}[t]{0.49\textwidth}
\centering
\renewcommand*{\arraystretch}{1.25}
\setlength{\tabcolsep}{3pt}
\resizebox{\columnwidth}{!}{
\begin{tabular}{l p{0em} ccs p{0.15em} ccs p{0.15em} ccs p{0em}}
\toprule
 \shortstack{\textcolor{darkgray}{\textbf{RxR}}\\\vphantom{\scriptsize a}}&& \multicolumn{3}{c}{\shortstack{Train\\ \scriptsize{($\subset$ \texttt{R2R train})}}} & &\multicolumn{3}{c}{\shortstack{Validation\\ \scriptsize{($\subset$ \texttt{R2R train})}}} && \multicolumn{3}{c}{\shortstack{Test\\ \scriptsize{($\subset$ \texttt{R2R val-seen})}}} & \\ \cmidrule(lr){3-5} \cmidrule(lr){7-9} \cmidrule(lr){11-13}
\multicolumn{2}{r}{Attacked:}        & {\cmark} & {\xmark} & {\large$\mathbf{\Delta}$} && {\cmark} & {\xmark} & {\large$\mathbf{\Delta}$} && {\cmark} & {\xmark} & {\large$\mathbf{\Delta}$} \\ \midrule

SR \textuparrow     && 24.47 & 63.17 & -38.70 && 29.86 & 62.90 & -33.04 && 25.59 & 59.06 & -33.47 \\
nDTW \textuparrow   && 27.17 & 66.94 & -39.77 && 32.48 & 66.91 & -34.43 && 29.28 & 62.67 & -33.39 \\ 

\bottomrule
\end{tabular}}%
\end{subtable}

\caption{\emph{Impact on VLN Performance for R2R (left) and RxR (right).} We report the impact on VLN performance for the HAMT agent encountering trajectory-level attacks in attacked (\cmark) and unaltered (\xmark) environments as well as the difference ($\mathbf{\Delta}$). 
Reductions in performance resulting in negative differences indicate our attacks are successful at disrupting the VLN agent's ability to follow given instructions.}
\label{tab:vln_with_without_attack}
\vspace{-6pt}
\end{table*}

\csubsection{Trajectory-level Attacks} 

We show trajectory-level attack results in \tabref{tab:attack_to_loc}. Overall, we find strong evidence that the attacks can induce multi-step trajectory following behaviors. 

For R2R (left), the HAMT agent operating attacked environments (\cmark) is roughly 5x more likely to arrive at the terminal location of the attack trajectory than in unaltered environments (\xmark) -- success rate climbing from 3.66\% to 21.61\% in \texttt{Test}. Likewise, nDTW trends that suggest the agent is more likely to adhere to the attack trajectory. We emphasize that examples in \texttt{Test} do not contribute to attack optimization or the training of the VLN agent. Each of these successes represent an attack generalizing to new natural language and observation sequences by correctly inducing the desired multi-step behavior in the multimodal sequential decision making policy of HAMT. Example attack objects are shown in \figref{fig:examples}.

For RxR (right), we find attacked agents are roughly 3x more likely to arrive at the terminal location of the attack trajectories, with \texttt{Test} success increasing by 15 points compared to unaltered environments. As in the previous experiments, absolute change in SR and nDTW are lower in RxR than R2R. To examine this, we also include oracle success rate (OSR) which checks if the agent arrives at the target location at any point in its trajectory -- essentially ignoring the \texttt{STOP} component of navigation. With OSR, we see a significant effect for RxR. Similar as Single-step case, we find the guiding trajectory followed by the attack trajectory averages 2-3 steps shorter than standard RxR training instances. This again suggests that while the attack is successful at redirecting the agent along the attack trajectory, the strong model bias for certain path lengths may be difficult to overcome. Further, RxR's broader set of instructions and paths may result in a more robust agent in terms adhering to the instruction -- as discussed in its proposal \cite{ku2020room}.

\xhdr{Effect on VLN Performance.}
Beyond inducing the desired trajectory, our trajectory-level attacks also significantly disrupt the agent's ability to follow its original instructions. In \tabref{tab:vln_with_without_attack}, we show standard VLN performance for HAMT in attacked (\cmark) and unaltered (\xmark) environments. We find our attacks cut success rate by nearly 35\% and 57\% in \texttt{Test} instances for R2R (left) and RxR (right) respectively with similar drops in other metrics. As in our prior results, these effects are stronger for samples from \texttt{Train} and \texttt{Validation} that are involved in the training of the attack. Note that performance in unaltered environments for \texttt{Test} does not match that reported in \tabref{tab:mesh-effect on vln} because only generated attack instances (\secref{sec:gen_attacks}) are considered here rather than all of \texttt{R2R-val-seen} / \texttt{RxR-val-seen-guide}.

\begin{table*}[t]
\begin{center}
\begin{subtable}[t]{0.24\textwidth}
\resizebox{\columnwidth}{!}{
\renewcommand*{\arraystretch}{1.3}
\setlength{\tabcolsep}{2pt}
\begin{tabular}{l p{0.5em} cc p{0.0em} cc p{0.0em}}
\toprule
 \shortstack{\footnotesize\textcolor{darkgray}{\textbf{R2R}}\\\vphantom{\scriptsize a}}&& \multicolumn{2}{c}{\shortstack{Validation\\ \scriptsize{($\subset$ \texttt{R2R train})}}} && \multicolumn{2}{c}{\shortstack{Test\\ \scriptsize{($\subset$ \texttt{R2R val-seen})}}} \\ \cmidrule(lr){3-4} \cmidrule(lr){6-7}

\multicolumn{1}{l}{\footnotesize Steps}  && \footnotesize SR \textuparrow &\footnotesize nDTW \textuparrow  &&\footnotesize SR \textuparrow & \footnotesize nDTW \textuparrow  \\ \midrule

1     && 6.86 & 41.18 && 4.08  & 33.17 \\
2     && 17.97 & 47.57 && 12.25  & 42.76\\
$3^*$   && \textbf{21.57} & \textbf{49.09} && \textbf{19.05} & \textbf{46.49} \\ \bottomrule
\end{tabular}}%
\caption{Steps Rendered}
\end{subtable}\hspace{0.0075\textwidth}
\begin{subtable}[t]{0.24\textwidth}
\resizebox{\columnwidth}{!}{
\renewcommand*{\arraystretch}{1.3}
\setlength{\tabcolsep}{2pt}
\begin{tabular}{l p{0.5em} cc p{0.0em} cc p{0.0em}}
\toprule
 \shortstack{\footnotesize\textcolor{darkgray}{\textbf{R2R}}\\\vphantom{\scriptsize a}}&& \multicolumn{2}{c}{\shortstack{Validation\\ \scriptsize{($\subset$ \texttt{R2R train})}}} && \multicolumn{2}{c}{\shortstack{Test\\ \scriptsize{($\subset$ \texttt{R2R val-seen})}}} \\ \cmidrule(lr){3-4} \cmidrule(lr){6-7}

\multicolumn{1}{l}{$\epsilon$}   && \footnotesize SR \textuparrow &\footnotesize nDTW \textuparrow  &&\footnotesize SR \textuparrow & \footnotesize nDTW \textuparrow  \\ \midrule

0.1     && 17.97& 46.83&& 16.33  & 42.57 \\
$0.3^*$     && 21.57 & 49.09 && 19.05  & 46.49 \\
0.5   && \textbf{21.90} & \textbf{49.68} && \textbf{22.45} & \textbf{46.55} \\ \bottomrule
\end{tabular}}
\caption{Attack Budget}
\end{subtable}\hspace{0.0075\textwidth}
\begin{subtable}[t]{0.24\textwidth}
\resizebox{\columnwidth}{!}{
\renewcommand*{\arraystretch}{1.3}
\setlength{\tabcolsep}{2pt}
\begin{tabular}{l p{0.5em} cc p{0.0em} cc p{0.0em}}
\toprule
 \shortstack{\footnotesize\textcolor{darkgray}{\textbf{R2R}}\\\vphantom{\scriptsize a}}&& \multicolumn{2}{c}{\shortstack{Validation\\ \scriptsize{($\subset$ \texttt{R2R train})}}} && \multicolumn{2}{c}{\shortstack{Test\\ \scriptsize{($\subset$ \texttt{R2R val-seen})}}} \\ \cmidrule(lr){3-4} \cmidrule(lr){6-7}

\multicolumn{1}{l}{\footnotesize Instr.}   && \footnotesize SR \textuparrow &\footnotesize nDTW \textuparrow  &&\footnotesize SR \textuparrow & \footnotesize nDTW \textuparrow  \\ \midrule

1     && 18.63 & 46.73 && 12.93  & 40.84 \\
2     && \textbf{21.90} & \textbf{49.37} && 17.69  & 42.95 \\
$3^*$   && 21.57 &  49.09 && \textbf{19.05} & \textbf{46.49} \\ \bottomrule
\end{tabular}}%
\caption{Instructions}
\end{subtable}\hspace{0.0075\textwidth}
\begin{subtable}[t]{0.24\textwidth}
\resizebox{\columnwidth}{!}{
\renewcommand*{\arraystretch}{1.35}
\setlength{\tabcolsep}{2pt}
\begin{tabular}{l p{0.5em} cc p{0.0em} cc p{0.0em}}
\toprule
 \shortstack{\footnotesize\textcolor{darkgray}{\textbf{R2R}}\\\vphantom{\scriptsize a}}&& \multicolumn{2}{c}{\shortstack{Validation\\ \scriptsize{($\subset$ \texttt{R2R train})}}} && \multicolumn{2}{c}{\shortstack{Test\\ \scriptsize{($\subset$ \texttt{R2R val-seen})}}} \\ \cmidrule(lr){3-4} \cmidrule(lr){6-7}

\multicolumn{1}{l}{\footnotesize Iters}   && \footnotesize SR \textuparrow &\footnotesize nDTW \textuparrow  &&\footnotesize SR \textuparrow & \footnotesize nDTW \textuparrow  \\ \midrule

$300^*$     && 21.57 & 49.09 && 19.05  & 46.49 \\
600     && 30.39 & 55.43 && 26.53  & 50.26 \\
900  && \textbf{31.70} & \textbf{56.31} && \textbf{26.53} & \textbf{51.71} \\ 
\bottomrule
\end{tabular}}
\caption{Training Iterations}
\end{subtable}
\end{center}
\vspace{-18pt}
\caption{\emph{Ablations of Attack Hyperparameters in R2R.} We use ${ }^*$ to denote baseline hyperparameters. During each ablation study, we vary only one hyperparameter at a time while use baseline hyperparameters for others.}
\label{tab:ablations_all}
\vspace{-6pt}
\end{table*}

\csubsection{Method Ablations}
To better understand the impact of hyperparameter choices on our attacks, we perform several ablations for trajectory-level attacks in R2R. Further details can be found in the supplementary materials. For space, we present results on \texttt{Validation} and \texttt{Test} instances only.

\xhdr{Steps Rendered.} %
For computational reasons, we only render observations during training for three viewpoints (see \secref{sec:method})-- the attack viewpoint and proceeding two viewpoints where the attack object is likely still visible. In \tabref{tab:ablations_all} (a), we vary the number of rendered observations during training between 1 (only the attack viewpoint) and 3. As in all our experiments, all observations are rendered during evaluation. We find rendering more steps improves attack performance significantly. For example, we observe a nearly 4x increase in \texttt{Test} success rate when rendering 3 steps instead of just 1. Rendering multiple observations better matches the evaluation setting and optimizes the attack to affect the agent across different viewing perspectives.

\xhdr{Attack Budget.} %
Varying $\epsilon$ allows for differing maximum object texture perturbations with larger $\epsilon$ denotes a stronger but more distorted and identifiable attack \cite{costa2024deep}. We use $\epsilon$'s of 0.1, 0.3, and 0.5 in \tabref{tab:ablations_all} (b). As expected, we find larger attack magnitudes lead to better attack performance.

\xhdr{Instructions.} %
Each trajectory in R2R is paired with 3 instructions. In \tabref{tab:ablations_all} (c), we vary how many of these are included during attack optimization per trajectory. Note we keep the training iterations the same across all runs. We find using more instructions during attack optimization generally leads to better generalization to the novel instructions encountered in \texttt{Validation} and \texttt{Test} instances.

\xhdr{Training Iterations.} %
Keeping batch size fixed at 16 and checkpoint selection on \texttt{Validation}, we vary the number of training iterations in \tabref{tab:ablations_all} (d). We find longer training times lead to substantial improvements in attack generalization -- e.g., increasing \texttt{Test} success rates by 7.5 points when training time is tripled from 300 to 900 iterations.

\csubsection{Factors that Influence Attack Success}

We investigate the effects from different factors on trajectory-level attack effectiveness on R2R \texttt{Test}. To facilitate our analysis, we frame experiments as paired-measurements on individual attack instances with nDTW measured pre- and post-attack. We assess the statistical significance of selected factors affecting nDTW by fitting linear mixed effect regression (\texttt{lmer}) models and evaluate significance with ANOVA and post-hoc t-tests.

\xhdr{Object Size.} We first examine effects of the attack object's size across navigation viewpoints. For each sub-image within each panoramic viewpoint, we calculate the attacked object's size as the percent of the sub-image covered by the object. For example, an object which cannot be seen will have a coverage of 0\% and one which fills the entire view 100\%. We find strong evidence ($p{=}0.005$) of a difference between pre/post-attack groups attributed to the size of the object in the selected view at the attack viewpoint. The coefficient of the effect indicates ($p{=}0.001$) a +0.59 point increase to attack nDTW per percent coverage. When considering the average coverage over all selected views in the attack trajectory, we again observe a significant difference between groups ($p{<}0.001$) and find the strength of the effect to increase to +1.50 ($p{<}0.001$) points per average percent covered. We note that the maximum observed coverage for a single selected view is 47\% and we would not suspect a linear trend in nDTW up to 100\% coverage. %

\xhdr{Object Category.} We analyze if any particular object category is easier or harder to attack and find strong evidence ($p{=}0.002$) for a difference between groups based on object category. Specifically, we find that the categories \texttt{sofa} (+15.15 nDTW, $p{=}0.028$) and \texttt{table} (+8.65 nDTW, $p{=}0.042$) are particularly susceptible to attack. Congruent with the size analysis above, we hypothesize that this is due at least somewhat to these objects being larger.%

\xhdr{Training Episode Diversity.} We find some evidence of a difference between groups ($p{=}0.081$) attributed to heading entropy when entering the attack viewpoint during training. We calculate heading entropy over the normalized rates of discrete entrance angles for training episodes entering the attack viewpoint. We find a positive correlation between entropy and nDTW (+8.41 nDTW/unit, $p{=}0.036$) indicating that greater trajectory diversity may improve performance. %

\xhdr{Object in Instruction.} Finally, we test for an effect from the attacked object being mentioned in the instruction. We use Porter stemming~\cite{porter1980stem} to reduce instruction words to their root form and then check for overlap in the WordNet \cite{miller-1994-wordnet} synsets of those words and the attacked objects class. We do not find any significant ($p{=}0.467$) effects. This observation is potentially supported by prior work that suggests object grounding is weak in VLN agents \cite{yang2023behavioral}.

\csection{Discussion}
\label{sec:conl}
In this work, we developed an adversarial attack framework that uses differentiable rendering to modify the appearance of 3D scene objects and demonstrated that it can allow an adversary to significantly impact the performance of VLN agents and even exert some level of control over their navigation trajectories regardless of the original user's instructions. At current, our work does not address the deployability of these attacks in the real world; however, prior work in adversarial machine learning \cite{kurakin2017adversarial, jan2019connecting} and sim2real transfer of VLN agents \cite{anderson2021sim} suggest it may be possible in future work. For now, our attacks also require substantial computation to produce due to the demands of differentiable rendering and assume a whitebox setting for the adversary; however, developing a blackbox analog is plausible at an increased computational expense. Together, this suggests that adequate adversarial defenses should be established prior to deployment of instructable embodied agents.

{\small
\bibliographystyle{ieee_fullname}
\bibliography{egbib}

\begin{thebibliography}{10}\itemsep=-1pt

\bibitem{abou2020systematic}
Anas Abou~Allaban, Maozhen Wang, and Ta{\c{s}}k{\i}n Pad{\i}r.
\newblock A systematic review of robotics research in support of in-home care for older adults.
\newblock {\em Information}, 11(2):75, 2020.

\bibitem{anderson2021sim}
Peter Anderson, Ayush Shrivastava, Joanne Truong, Arjun Majumdar, Devi Parikh, Dhruv Batra, and Stefan Lee.
\newblock Sim-to-real transfer for vision-and-language navigation.
\newblock In {\em Conference on Robot Learning}, pages 671--681. PMLR, 2021.

\bibitem{anderson2018vision}
Peter Anderson, Qi Wu, Damien Teney, Jake Bruce, Mark Johnson, Niko Sünderhauf, Ian Reid, Stephen Gould, and Anton van~den Hengel.
\newblock Vision-and-language navigation: Interpreting visually-grounded navigation instructions in real environments.
\newblock {\em CVPR}, 2018.

\bibitem{athalye2017synthesizing}
Anish Athalye, Logan Engstrom, Andrew Ilyas, and Kevin Kwok.
\newblock Synthesizing robust adversarial examples.
\newblock {\em ICML}, 2017.

\bibitem{barreno2006can}
Marco Barreno, Blaine Nelson, Russell Sears, Anthony~D. Joseph, and J.~D. Tygar.
\newblock Can machine learning be secure?
\newblock In {\em Proceedings of the 2006 ACM Symposium on Information, Computer and Communications Security}, ASIACCS '06, page 16–25, New York, NY, USA, 2006. Association for Computing Machinery.

\bibitem{brown2017adversarial}
Tom~B. Brown, Dandelion Mané, Aurko Roy, Martín Abadi, and Justin Gilmer.
\newblock Adversarial patch.
\newblock {\em arXiv}, 2017.

\bibitem{chang2017matterport3d}
Angel Chang, Angela Dai, Thomas Funkhouser, Maciej Halber, Matthias Niebner, Manolis Savva, Shuran Song, Andy Zeng, and Yinda Zhang.
\newblock Matterport3d: Learning from rgb-d data in indoor environments.
\newblock In {\em 2017 International Conference on 3D Vision (3DV)}, pages 667--676. IEEE, 2017.

\bibitem{chen2021history}
Shizhe Chen, Pierre-Louis Guhur, Cordelia Schmid, and Ivan Laptev.
\newblock History aware multimodal transformer for vision-and-language navigation.
\newblock In {\em NeurIPS}, 2021.

\bibitem{costa2024deep}
Joana~C Costa, Tiago Roxo, Hugo Proen{\c{c}}a, and Pedro~RM In{\'a}cio.
\newblock How deep learning sees the world: A survey on adversarial attacks \& defenses.
\newblock {\em IEEE Access}, 2024.

\bibitem{embodiedqa}
Abhishek Das, Samyak Datta, Georgia Gkioxari, Stefan Lee, Devi Parikh, and Dhruv Batra.
\newblock {E}mbodied {Q}uestion {A}nswering.
\newblock In {\em Proceedings of the IEEE Conference on Computer Vision and Pattern Recognition (CVPR)}, 2018.

\bibitem{dosovitskiy2020image}
Alexey Dosovitskiy.
\newblock An image is worth 16x16 words: Transformers for image recognition at scale.
\newblock {\em arXiv preprint arXiv:2010.11929}, 2020.

\bibitem{duan2020adversarial}
Ranjie Duan, Xingjun Ma, Yisen Wang, James Bailey, A~Kai Qin, and Yun Yang.
\newblock Adversarial camouflage: Hiding physical-world attacks with natural styles.
\newblock In {\em CVPR}, 2020.

\bibitem{eykholt2018robust}
Kevin Eykholt, Ivan Evtimov, Earlence Fernandes, Bo Li, Amir Rahmati, Chaowei Xiao, Atul Prakash, Tadayoshi Kohno, and Dawn Song.
\newblock Robust physical-world attacks on deep learning visual classification.
\newblock In {\em CVPR}, 2018.

\bibitem{gu2022vision}
Jing Gu, Eliana Stefani, Qi Wu, Jesse Thomason, and Xin Wang.
\newblock Vision-and-language navigation: A survey of tasks, methods, and future directions.
\newblock {\em ACL}, 2022.

\bibitem{huang2017adversarial}
Sandy~H. Huang, Nicolas Papernot, Ian~J. Goodfellow, Yan Duan, and P. Abbeel.
\newblock Adversarial attacks on neural network policies.
\newblock {\em ICLR (Workshop)}, 2017.

\bibitem{ilharco2019general}
Gabriel Ilharco, Vihan Jain, Alexander Ku, Eugene Ie, and Jason Baldridge.
\newblock General evaluation for instruction conditioned navigation using dynamic time warping.
\newblock {\em arXiv preprint arXiv:1907.05446}, 2019.

\bibitem{jan2019connecting}
Steve~T.K. Jan, Joseph Messou, Yen-Chen Lin, Jia-Bin Huang, and Gang Wang.
\newblock Connecting the digital and physical world: Improving the robustness of adversarial attacks.
\newblock In {\em AAAI}, 2019.

\bibitem{kingma2014adam}
Diederik~P Kingma and Jimmy Ba.
\newblock Adam: A method for stochastic optimization.
\newblock {\em arXiv preprint arXiv:1412.6980}, 2014.

\bibitem{ku2020room}
Alexander Ku, Peter Anderson, Roma Patel, Eugene Ie, and Jason Baldridge.
\newblock Room-across-room: Multilingual vision-and-language navigation with dense spatiotemporal grounding.
\newblock {\em arXiv preprint arXiv:2010.07954}, 2020.

\bibitem{kurakin2017adversarial}
Alexey Kurakin, Ian~J. Goodfellow, and Samy Bengio.
\newblock Adversarial examples in the physical world.
\newblock {\em ICLR}, 2017.

\bibitem{li2019adversarial}
Juncheng Li, Frank~R. Schmidt, and J.~Zico Kolter.
\newblock Adversarial camera stickers: {A} physical camera-based attack on deep learning systems.
\newblock {\em ICML}, 2019.

\bibitem{lin2017tactics}
Yen-Chen Lin, Zhang-Wei Hong, Yuan-Hong Liao, Meng-Li Shih, Ming-Yu Liu, and Min Sun.
\newblock Tactics of adversarial attack on deep reinforcement learning agents.
\newblock {\em IJCAI}, 2017.

\bibitem{liu2020spatiotemporal}
Aishan Liu, Tairan Huang, Xianglong Liu, Yitao Xu, Yuqing Ma, Xinyun Chen, Stephen~J. Maybank, and Dacheng Tao.
\newblock Spatiotemporal attacks for embodied agents.
\newblock {\em ECCV}, 2020.

\bibitem{luo2024imageworth1000lies}
Haochen Luo, Jindong Gu, Fengyuan Liu, and Philip Torr.
\newblock An image is worth 1000 lies: Adversarial transferability across prompts on vision-language models, 2024.

\bibitem{madry2018towards}
Aleksander Madry, Aleksandar Makelov, Ludwig Schmidt, Dimitris Tsipras, and Adrian Vladu.
\newblock Towards deep learning models resistant to adversarial attacks.
\newblock {\em ICLR}, 2018.

\bibitem{miao2022isometric}
Yibo Miao, Yinpeng Dong, Junyi Zhu, and Xiao-Shan Gao.
\newblock Isometric 3d adversarial examples in the physical world.
\newblock {\em NeurIPS}, 2022.

\bibitem{miller-1994-wordnet}
George~A. Miller.
\newblock {W}ord{N}et: A lexical database for {E}nglish.
\newblock In {\em {H}uman {L}anguage {T}echnology: Proceedings of a Workshop held at {P}lainsboro, {N}ew {J}ersey, {M}arch 8-11, 1994}, 1994.

\bibitem{moosavi_dezfooli2016universal}
Seyed-Mohsen Moosavi-Dezfooli, Alhussein Fawzi, Omar Fawzi, and Pascal Frossard.
\newblock Universal adversarial perturbations.
\newblock {\em CVPR}, 2016.

\bibitem{morris2020textattack}
John~X Morris, Eli Lifland, Jin~Yong Yoo, Jake Grigsby, Di Jin, and Yanjun Qi.
\newblock Textattack: A framework for adversarial attacks, data augmentation, and adversarial training in nlp.
\newblock {\em arXiv preprint arXiv:2005.05909}, 2020.

\bibitem{pestana2022transferable}
Camilo Pestana, Naveed Akhtar, Nazanin Rahnavard, Mubarak Shah, and Ajmal~S. Mian.
\newblock Transferable 3d adversarial textures using end-to-end optimization.
\newblock {\em WACV}, 2022.

\bibitem{porter1980stem}
M.F. Porter.
\newblock An algorithm for suffix stripping.
\newblock In {\em Program}, pages 130--137, 1980.

\bibitem{qi2020reverie}
Yuankai Qi, Qi Wu, Peter Anderson, Xin Wang, William~Yang Wang, Chunhua Shen, and Anton van~den Hengel.
\newblock Reverie: Remote embodied visual referring expression in real indoor environments.
\newblock In {\em Proceedings of the IEEE/CVF Conference on Computer Vision and Pattern Recognition}, pages 9982--9991, 2020.

\bibitem{qiaoben2022consistent}
You Qiaoben, Chengyang Ying, Xinning Zhou, Hang Su, Jun Zhu, and Bo Zhang.
\newblock Consistent attack: Universal adversarial perturbation on embodied vision navigation.
\newblock {\em Pattern Recognition Letters}, 168, 2022.

\bibitem{qiu2022adversarial}
Shilin Qiu, Qihe Liu, Shijie Zhou, and Wen Huang.
\newblock Adversarial attack and defense technologies in natural language processing: A survey.
\newblock {\em Neurocomputing}, 492:278--307, 2022.

\bibitem{ravi2020accelerating}
Nikhila Ravi, Jeremy Reizenstein, David Novotny, Taylor Gordon, Wan-Yen Lo, Justin Johnson, and Georgia Gkioxari.
\newblock Accelerating 3d deep learning with pytorch3d.
\newblock {\em arXiv preprint arXiv:2007.08501}, 2020.

\bibitem{roth2021token}
Tom Roth, Yansong Gao, Alsharif Abuadbba, Surya Nepal, and Wei Liu.
\newblock Token-modification adversarial attacks for natural language processing: A survey.
\newblock {\em AI Communications}, (Preprint):1--22, 2021.

\bibitem{sodiya2024ai}
Enoch~Oluwademilade Sodiya, Uchenna~Joseph Umoga, Olukunle~Oladipupo Amoo, and Akoh Atadoga.
\newblock Ai-driven warehouse automation: A comprehensive review of systems.
\newblock {\em GSC Advanced Research and Reviews}, 18(2):272--282, 2024.

\bibitem{DBLP:journals/corr/SzegedyZSBEGF13}
Christian Szegedy, Wojciech Zaremba, Ilya Sutskever, Joan Bruna, Dumitru Erhan, Ian~J. Goodfellow, and Rob Fergus.
\newblock Intriguing properties of neural networks.
\newblock In Yoshua Bengio and Yann LeCun, editors, {\em 2nd International Conference on Learning Representations, {ICLR} 2014, Banff, AB, Canada, April 14-16, 2014, Conference Track Proceedings}, 2014.

\bibitem{wang2021dual}
Jiakai Wang, Aishan Liu, Zixin Yin, Shunchang Liu, Shiyu Tang, and Xianglong Liu.
\newblock Dual attention suppression attack: Generate adversarial camouflage in physical world.
\newblock {\em CVPR}, 2021.

\bibitem{wu2020making}
Zuxuan Wu, Ser-Nam Lim, Larry Davis, and Tom Goldstein.
\newblock Making an invisibility cloak: Real world adversarial attacks on object detectors.
\newblock {\em ECCV}, 2020.

\bibitem{yang2023behavioral}
Zijiao Yang, Arjun Majumdar, and Stefan Lee.
\newblock Behavioral analysis of vision-and-language navigation agents.
\newblock In {\em Proceedings of the IEEE/CVF Conference on Computer Vision and Pattern Recognition}, pages 2574--2582, 2023.

\bibitem{yoo2021towards}
Jin~Yong Yoo and Yanjun Qi.
\newblock Towards improving adversarial training of nlp models.
\newblock {\em arXiv preprint arXiv:2109.00544}, 2021.

\bibitem{zhang2020robust}
Huan Zhang, Hongge Chen, Chaowei Xiao, Bo Li, Mingyan~D. Liu, Duane~S. Boning, and Cho-Jui Hsieh.
\newblock Robust deep reinforcement learning against adversarial perturbations on state observations.
\newblock {\em NeurIPS}, 2020.

\bibitem{zhang20213d}
Jinlai Zhang, Lyujie Chen, Binbin Liu, Bojun Ouyang, Qizhi Xie, Jihong Zhu, and Yanmei Meng.
\newblock 3d adversarial attacks beyond point cloud.
\newblock {\em Information Sciences}, 2021.

\bibitem{zhang2022navigation}
Yunchao Zhang, Zonglin Di, Kaiwen Zhou, Cihang Xie, and Xin~Eric Wang.
\newblock Navigation as the attacker wishes? towards building byzantine-robust embodied agents under federated learning.
\newblock {\em arXiv}, 2022.

\bibitem{zhao2023evaluate}
Yunqing Zhao, Tianyu Pang, Chao Du, Xiao Yang, Chongxuan Li, Ngai-Man Cheung, and Min Lin.
\newblock On evaluating adversarial robustness of large vision-language models.
\newblock In {\em Thirty-seventh Conference on Neural Information Processing Systems}, 2023.

\end{thebibliography}
}
\newpage
\appendix

\section{Data Description}
\subsection{R2R Data}
As described in Sec 3.2, we select attack instances from \texttt{R2R-val-seen} that have sufficient support from \texttt{R2R-train}. In total, we generated 1735 episodes as \texttt{Train}, 577 episodes as \texttt{Validation} that corresponds to 273 attack instances as \texttt{Test}. The attack instances cover 68 unique objects spanning 39 environments. Notice the attack for each attack instance trains independently, so each attack instance has their corresponding train, validation set. Here we provide the aggregated number.
\subsection{RxR Data}
As attacks on RxR data taking substantially more compute and time compared to R2R, we random sample a subset based on the number of unique attack objects involved, which results in a subset covering  20 unique objects residing in 9 environments. As a result, we have 1659 episodes for \texttt{Train}, 345 for \texttt{Validation} that corresponds to 254 attack instances in \texttt{Test}. This is comparable to the number of attack instances of R2R.
\subsection{Ablation Data}
Similarly, to accommodate for time and compute constraints, we random sample a subset from R2R, that covers 34 unique objects out of 68 in total, which spans across 27 environments and result in 955 \texttt{Train}, 306 \texttt{Validation} and 147 \texttt{Test} attack instances, that is roughly half of the total dataset on which we reported main result.

\section{lmer Construction for Factor Analysis}
We investigate the effects from different factors on trajectory-level attack effectiveness on R2R \texttt{Test}. To facilitate our analysis, we frame experiments as paired-measurements on individual attack instances with nDTW measured pre- and post-attack. Let $Y_{hijk}$ be the response variable nDTW, where $j,k$ respectively index random effect grouping factors for individual objects \smash{$object \sim N(0,\sigma^2_o)$} and attack instances \smash{$instance \sim N(0,\sigma^2_s)$}. We assess the statistical significance of some $predictor$ (e.g., object size) indexed by $i$ with $n$ levels affecting nDTW by fitting linear mixed effect regression (\texttt{lmer}) models of the form:

\begin{multline}
    Y_{hijk} = 
    \beta_0 + 
    \beta_{1}attack_h + 
    \sum_{m=2}^{n}\beta_{m}(attack_{h} \times predictor_{i})_m \\
    + object_j + 
    instance_{k} + 
    \epsilon_{hijk},
\end{multline}

where $\epsilon_{hijk} \sim N(0,1)$ is a random error term, $\beta_0$ the model intercept, $\beta_1$ the fixed effect of $attack$, and $\beta_{2:n}$ the fixed effect of the interactions between $attack$ and $predictor$. Note that there are exactly two samples for each $instance_{k}$, one for pre-attack ($attack_{h{=}0}$) and one for post-attack ($attack_{h{=}1}$). Informally, we model paired nDTW measurements as a main effect from applying the adversarial attack, the interaction effect between the attack and some predictor, and random intercept effects from attack instances and objects. We use an ANOVA to determine the overall significance of factors and examine graphical model residual diagnostics to validate its modeling assumptions. For factors with significant effects, we use a post-hoc t-test to determine if the coefficients relating to post-attack $predictor$ interactions are significantly different from zero. That is, we verify that the difference in effect from the $predictor$ in pre/post-attack measurements is significant, and that the strength of that effect in the post-attack setting is significantly different from zero.

\end{document}